%% file: acl2023.tex
\documentclass[11pt]{article}

\usepackage[]{Format/acl2023}
\usepackage{balance}

\usepackage{times}
\usepackage{booktabs}
\usepackage{tikz}
\usetikzlibrary{shapes,arrows,positioning}

\usepackage{hyperref}
\usepackage{latexsym}
\usepackage{multirow}
\usepackage[T1]{fontenc}
\usepackage[utf8]{inputenc}

\usepackage{microtype}

\usepackage{graphicx}
\usepackage{multirow}
\usepackage{amsmath}
\usepackage{amssymb}

\usepackage{CJKutf8}
\usepackage{color, colortbl}
\usepackage{soul}

\title{Learning-From-Mistakes Prompting for Indigenous Language Translation}

\author{\textbf{You-Cheng Liao}, \textbf{Chen-Jui Yu}, \textbf{Chi-Yi Lin},  \textbf{He-Feng Yun}, \\ \textbf{Yen-Hsiang Wang}, \textbf{Hsiao-Min Li}, \textbf{Yao-Chung Fan$^*$}\\
  Department of Computer Science and Engineering, \\
  National Chung Hsing University, Taiwan\\
  \texttt{yfan@nchu.edu.tw}
  }

\begin{document} \begin{CJK*}{UTF8}{bsmi}

\maketitle
\begin{abstract}
Using large language models, this paper presents techniques to improve extremely low-resourced indigenous language translations. Our approaches are grounded in the use of (1) the presence of a datastore consisting of a limited number of parallel translation examples, (2) the inherent capabilities of LLMs like GPT-3.5, and (3) a word-level translation dictionary. We harness the potential of LLMs and in-context learning techniques in such a setting for using LLM as universal translators for extremely low-resourced languages. Our methodology hinge on utilizing LLMs as language compilers for selected language pairs, hypothesizing that they could internalize syntactic structures to facilitate accurate translation. We introduce three techniques: \textit{KNN-Prompting with Retrieved Prompting Context, Chain-of-Thought Prompting, and Learning-from-Mistakes Prompting}, with the last method addressing past errors. The evaluation results suggest that, even with limited corpora, LLMs, when paired with proper prompting, can effectively translate extremely low-resource languages.
\end{abstract}

\section{Introduction}
% Taiwan is renowned for its array of indigenous languages, which reflect its rich cultural heritage. However, these languages are diminishing due to urbanization and the dominance of Mandarin Chinese \cite{pawan2009indigenous} Recently, efforts to preserve these languages have intensified, recognizing their role in fostering cultural diversity and strengthening indigenous identities \cite{lin2007indigenous}. As a result, there's a growing need to bridge these languages with more prevalent ones like Mandarin Chinese.

In recent years, LLMs have showcased astonishing capabilities in the realm of natural language processing, particularly in tasks like language translation \cite{zhu2023multilingual}, text generation \cite{yuan2022selecting}, and contextual understanding \cite{behnia2022ew}. The robust functionality of these models has led us to reconsider their potential role in indigenous language translation. 
% Especially for Taiwan's indigenous languages, which face considerable challenges due to resource constraints, the intervention of LLMs offers a unique opportunity. 

In our pursuit to facilitate translations from Chinese to Taiwanese indigenous languages, we leverage the power of LLMs, buttressed by three foundational pillars: the presence of a datastore consisting of a limited number of parallel translation examples, the inherent capabilities of LLMs like GPT-3.5, and the integration of a word-level translation dictionary. 

\begin{figure*}[t]
    \centering
    \includegraphics[width=.9\linewidth]{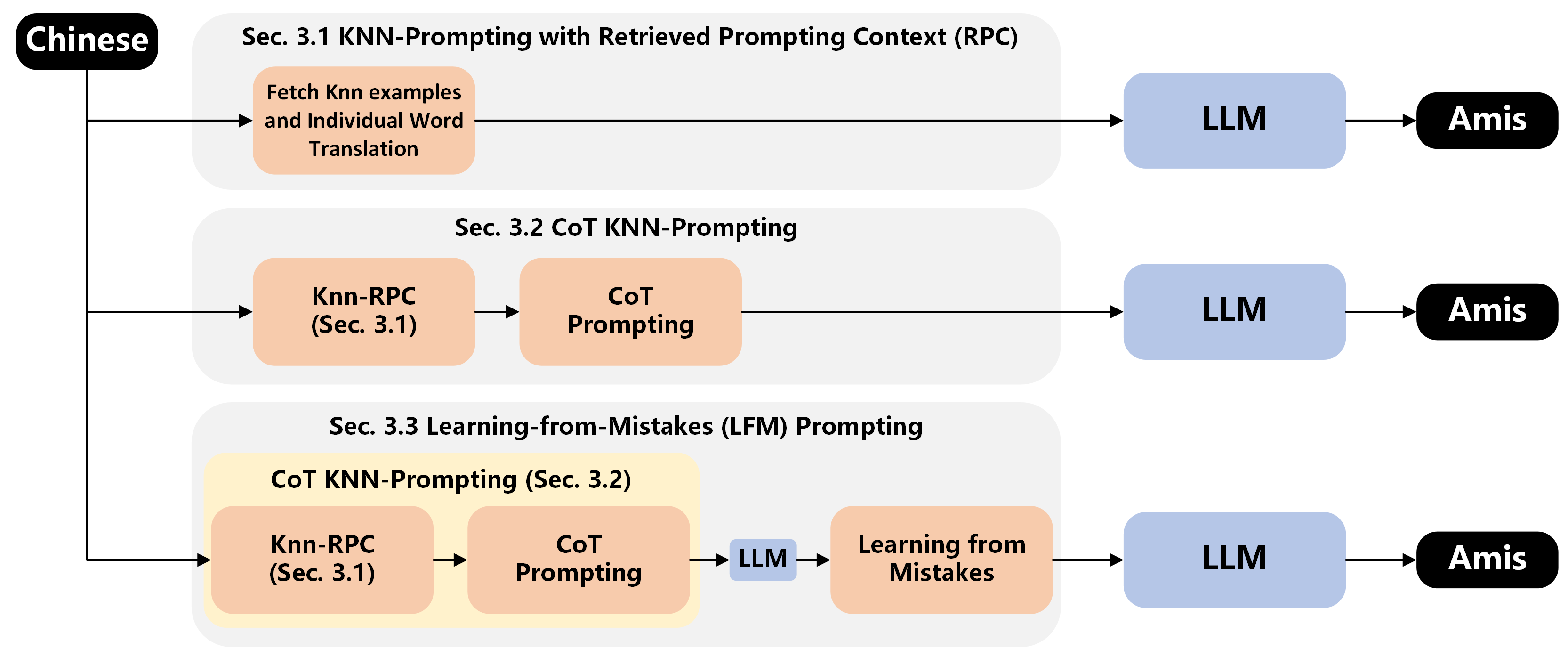}
    \caption{Methodology Overview}
    \label{fig:methodolgy-overview}
\end{figure*}

In this paper, we delineate three translation methodologies that build upon each other in a cumulative fashion. Each method represents a layer in our stratified approach, starting from leveraging contextual similarity in KNN-Prompting with Retrieved Prompting Context (RPC) to harnessing the didactic potential of Chain of Thought (CoT) Prompting, and culminating in the Learning-from-Mistakes (LFM) Prompting technique that incorporates feedback mechanisms for continuous improvement. Figure \ref{fig:methodolgy-overview} provides an overview of our methodologies, illustrating a step-by-step translation enhancement process designed for the Taiwanese indigenous language context.

This paper is structured as follows: In Section \ref{sec:rel}, we review the literature and discuss the position of this study. In Section \ref{sec:method}, we explore the CoT Prompting methodology, followed by an in-depth analysis of the LFM Prompting approach. In Section \ref{sec:experiment}, we report the evaluation results. Through empirical evaluation and expert reviews, we demonstrate the effectiveness of the proposed methodologies. 

% This study unfolds three translation methodologies.

% \begin{itemize}
% \item \textbf{KNN-Prompting:} Grounded in few-shot learning principles \cite{brown2020language}, this technique utilizes the $k$-nearest neighbor algorithm to provide the LLM with contextually relevant prompts sourced from a parallel corpus and a word translation dictionary. It pinpoints the most contextually pertinent examples in the datastore to anchor the translation.

% \item \textbf{CoT Prompting:} An evolution of KNN-Prompting, CoT Prompting equips the LLM with a step-by-step understanding of the target language's syntax and sentence construction, guiding the model through a logical sequence of word translations and sentence alignment.

% \item \textbf{Learning-from-Mistakes (LFM) Prompting:} This two-stage approach begins with a preliminary translation using CoT Prompting. It then leverages a comparison with previous translation mistakes, instructing the LLM to correct errors and enhance the overall translation accuracy.
% \end{itemize}

% Together, these methods form a comprehensive strategy to surmount the challenges faced when translating into resource-scarce indigenous languages. They demonstrate the untapped potential of LLMs in bridging linguistic divides and enriching the communication within and across indigenous communities.

\section{Related Work}\label{sec:rel}
LLMs have exhibited excellent performance in language translation tasks, particularly evident in well-represented source languages like English and Chinese. Despite significant strides in translation performance for these languages, there remains a notable gap in the exploration of LLMs for low-resourced languages or those that have not been pre-trained. This aspect represents an under-explored area within the domain of research.

\subsection{Low Resource Translation with LLM}
% llm的效能兩個主要因素：in-context learning & instruction ability 
LLMs' effectiveness on various task is primarily attributed to two main properties. Firstly, in-context learning \cite{brown2020language, lester2021power} allows the model to learn to solve specific problems by providing a small number of examples within the input context. The second one is the ability to follow the instruction \cite{ouyang2022training,mishra2021cross,wei2021finetuned}, instruction-tuned LLMs can be guiding to solve new task based on text instruction just as the scenario they were trained.  

% 提及LLM做low resource的work
Recently, some research has focused on enhancing these instruction following LLMs through in-context learning, \cite{nguyen2023democratizing,ahuja2023mega} explores the generation of unsupervised few-shot demonstrations to enhance translation effectiveness in low-resource scenarios. 
Additionally, \cite{yao2023empowering} utilized cultural awareness to optimize alignment in different languages, further augmenting the translation performance of LLMs.

%大多數low resource與我們的差異,1. 是否存在llm pretrain的corpus 2. 有別於以往用訓練參數得手法
It is noteworthy that, the mentioned works above have focused on low-resource data for LLMs in languages that have been encountered during the pre-training phase. 
In contrast, our emphasis lies in a scenario where the model has not been previously trained in this specific language.
In contrast to conventional approaches, we refrain from training parameters on limited parallel corpora \cite{gu2018universal,lalrempuii2023investigating}. Instead, our goal is to leverage the understanding and reasoning capabilities of LLMs, coupled with the provided data, to accomplish translation tasks for previously unseen languages.

In summary, to the best of our knowledge, no study has delved into the challenges and applications of utilizing LLMs for languages that have not been encountered before.

\subsection{Indigenous Language Translation}
In the context of preserving and revitalizing indigenous languages, the work by \cite{zheng2022parallel} stands as a notable contribution. Zheng and colleagues introduce the Amis-Mandarin dataset, which includes a parallel corpus comprising 5,751 Amis and Mandarin sentences. This dataset is of particular relevance to our research on translating Chinese sentences into Taiwanese indigenous languages. The Amis-Mandarin dataset provides a valuable resource for studying indigenous language translation. It aligns with the objectives of our study, as it offers a substantial parallel corpus, a fundamental component for training and evaluating translation models. Our research similarly leverages parallel corpora, although we focus on the translation of Chinese into various indigenous languages, including but not limited to Amis. In this study, we conduct experiments on six different indigenous languages. 

Furthermore, \cite{zheng2022parallel} compile a comprehensive dictionary containing 7,800 unique Amis words and phrases, each accompanied by its Mandarin definition. This lexical resource enhances the utility of their dataset for translation tasks. In our research, we assume the existence of a similar dictionary, emphasizing the importance of word-level translation between Chinese and Taiwanese indigenous languages.

\citet{stap2023chatgpt} evaluates the translation performance of different systems for Spanish to 11 indigenous languages from South America. The authors find that LLMs like ChatGPT are not yet good at translating into indigenous languages. This is likely due to a number of factors, including the lack of training data for indigenous languages, the complex grammar and sentence structure of indigenous languages, and the difficulty of capturing the nuances of indigenous culture in translation.

\subsection{Unveiling LLMs' Proficiency in Tool Usage}

In recent research, \cite{schick2023toolformer} discovered that LLMs exhibit the ability to discern how to employ tools provided by users, including external data. They adeptly combine this external information with their own knowledge to effectively address problem -solving tasks. These investigations delve into the mechanisms of CoT \cite{inaba2023multitool} and Self-instruction \cite{yang2023gpt4tools} approaches, exploring how these methodologies assist LLMs in comprehending questions and utilizing the tools at their disposal. Additionally, there has been the development of question- answering datasets, such as ToolQA \cite{zhuang2023toolqa,inaba2023multitool}, which aimed at faithfully evaluating the ability of LLMs to use external tools for question-answering.

Inspired by these explorations into the understanding and application capabilities of LLMs, we take a similar approach in our method design. We offer KNN examples and word-by-word translation as tools for LLMs to improve their language translation abilities.

 \subsection{Position of Our Paper}

This research stands at the intersection of multiple areas, addressing the challenges of translating into low-resource indigenous languages using LLMs like ChatGPT. While prior works have explored low-resource translation and indigenous language preservation, our study distinguishes itself in two key aspects:

\begin{enumerate}
    \item \textbf{Languages Unseen in Pre-training:} Unlike previous research that has primarily focused on low-resource data for LLMs in languages encountered during pre-training, our work emphasizes the scenario where the model has not been previously trained in the specific target language. We tackle the challenge of translating into languages that lack representation in the model's training data, making our approach more versatile and applicable to a broader range of indigenous languages.
    
    \item \textbf{Few-Shot Prompting Techniques:} Our research pioneers the application of few-shot prompting techniques to enhance translation capabilities for indigenous languages. We introduce innovative methods, including \textit{KNN-Prompting with RPC, CoT Prompting, and LFM Prompting}, tailored to leverage LLMs' inherent understanding and reasoning abilities. These techniques empower LLMs to effectively tackle low-resource language translation tasks, even when working with limited parallel corpora.
\end{enumerate}

In summary, our paper bridges the gap between LLMs and low-resource indigenous language translation, offering practical and innovative solutions for preserving and revitalizing endangered languages. By exploring the potential of these models in an uncharted linguistic landscape, we provide a fresh perspective and a promising direction for future research in this domain. For clarity, we also compare the related work in Table \ref{tab:overview}. 

%Table
\begin{table}[t]
\centering
\resizebox{\linewidth}{!}{
\begin{tabular}{|l|l|l|l|l|}
\hline
 paper& \begin{tabular}{c}
 In-context       \\
  Learning     
 \end{tabular}
    & \begin{tabular}{c}
  Fine-tune    \\
  Parame.     
 \end{tabular} &\begin{tabular}{c}
    Low-Resource      \\
    Language  
    \end{tabular}    & 
    \begin{tabular}{c}
    Unseen      \\
    Language  
    \end{tabular}
     \\ 
\hline
\cite{yao2023empowering} & \checkmark &  &  &   \\
\hline
\cite{nguyen2023democratizing} & \checkmark &  & \checkmark &   \\
\hline
\cite{guerreiro2023hallucinations} & \checkmark &  &\checkmark  &   \\
\hline
\cite{gu2018universal} &  &  \checkmark & \checkmark &  \checkmark  \\
\hline
\cite{lalrempuii2023investigating} &  &  \checkmark & \checkmark &  \checkmark  \\
\hline
Our work & \checkmark &  & \checkmark &  \checkmark \\
\hline
\end{tabular}}
\caption{An overview of the existing language translation studies}
\label{tab:overview}
\end{table}

% The paper \cite{zheng2022parallel} introduces a significant contribution to the field by presenting an Amis-Mandarin dataset, which comprises a parallel corpus of 5,751 Amis and Mandarin sentences, as well as a dictionary containing 7,800 unique Amis words and phrases with their Mandarin definitions. In addition to this dataset, the authors also report the training of neural machine translation models on this dataset, establishing a foundation for further research in the area.

\section{Methodology}\label{sec:method}
\noindent\textbf{Problem Setting and Assumptions} The primary objective of this research is to enable the translation of Chinese sentences into Taiwanese indigenous languages through the utilization of LLMs. In pursuit of this goal, we make the following assumptions for the methods proposed in this study:

\begin{itemize} \item \textbf{Datastore of Parallel Corpora:} Our first assumption centers on the availability of a datastore with limited translation examples. Within this datastore, each data entry comprises a pair of sentences: a source sentence in Chinese (the language intended for translation) and a corresponding target sentence in the specific Taiwanese indigenous language. This resource forms the backbone for our translation, facilitating the alignment of linguistic patterns and meanings.
    
    \item \textbf{Large Language Models:} The cornerstone of our translation methods is the utilization of large pre-trained language models, exemplified by GPT-3.5, as the primary translation engines. 
    
    \item \textbf{Dictionary Existence:} In addition to the aforementioned resources, we introduce another assumption: the existence of a dictionary that spans word-level translations. This dictionary encompasses translations between indigenous language words and their corresponding Chinese counterparts. 
\end{itemize}

Figure \ref{fig:methodolgy-overview} outlines our study's methods for enhancing translation in a cumulative manner. The \textit{KNN-Prompting with RPC} method forms the base, merging contextually similar sentences and word translations to inform the LLM's understanding of grammar and context. The \textit{CoT Prompting} adds CoT demonstrations, showing RPC integration for effective translation. The \textit{LFM Prompting} expands upon these with a feedback loop, leveraging previous translation errors to refine outcomes. This progressive strategy not only enhances LLM's translation proficiency but also promotes continual learning and accuracy improvement.

\subsection{KNN-Prompting with Retrieved Prompting Context (RPC)}
We investigate the application of few-shot learning through the KNN-Prompting concept, as discussed in the works of Shi et al. \cite{shi2022nearest} and Xu et al. \cite{xu2023k}. Our approach not only leverages contextually similar examples but also incorporates individual translations for each word in the source language. The methodology unfolds in the following manner:

\begin{itemize}
    \item When tasked with translating a sentence $s$, our method initiates by constructing a \textit{Retrieved Prompting Context (RPC)} for $s$. This context includes:
    \begin{itemize}
        \item $k$ examples that are contextually analogous to $s$, selected based on their similarity.
        \item Translations for each word in $s$, sourced from a comprehensive dictionary.
    \end{itemize}

    \item For instances where direct word equivalents are unavailable, we employ the BERT-base-chinese model as an embedding tool. This model aids in computing similarities to identify the most appropriate substitute words.

    \item The core principle of our method is to enable the LLM to assimilate the grammatical norms and sentence constructs of the target language. It achieves this through the analysis of the $k$ examples, thereby learning to organize the individually translated words into coherent and grammatically consistent sentences.
\end{itemize}

\begin{figure}[]
\centering\includegraphics[width=\linewidth]{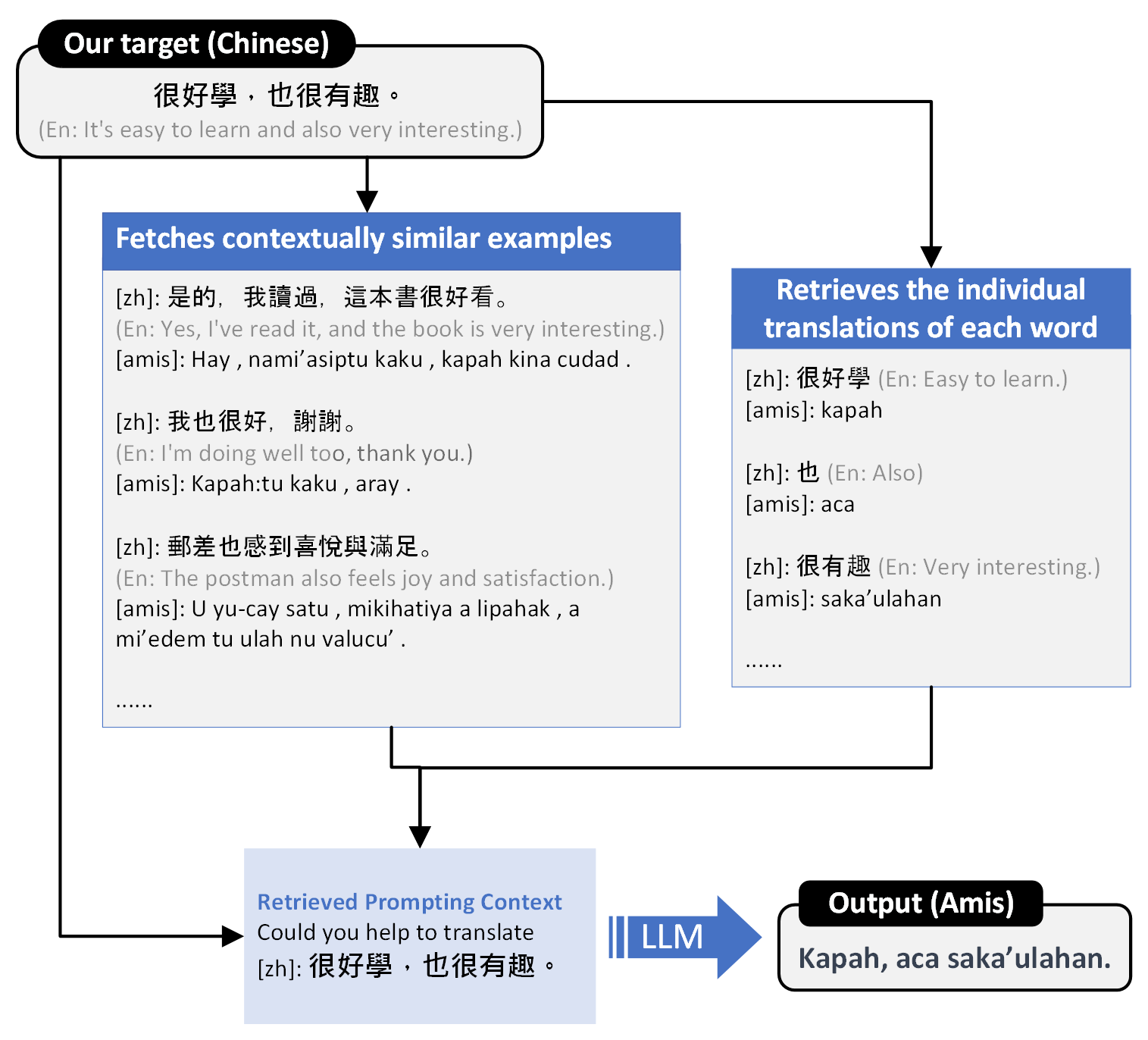}
    \caption{KNN-Prompting with RPC}
    \label{fig:KNN-Prompting-RPC}
\end{figure}

For a practical illustration of this process, please see Figure \ref{fig:KNN-Prompting-RPC}, which provides a concrete example of the RPC in action. We also show an example for prompting in Table \ref{tab:RPC}.

\begin{figure*}[t]
    \centering
    \includegraphics[width=.9\textwidth]{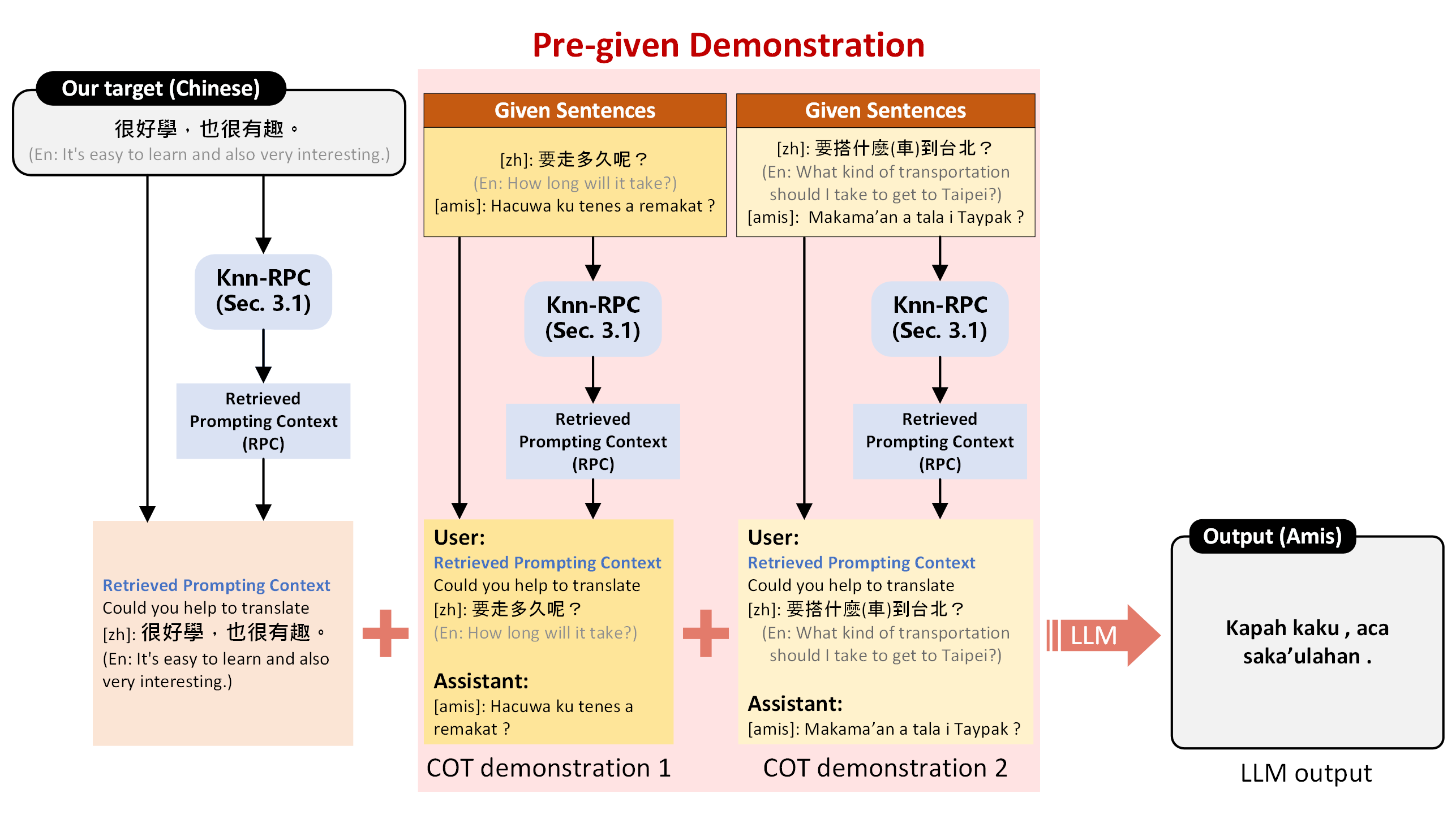}
    \caption{CoT KNN-Prompting: In this example, we have two CoT demonstrations. Note that each CoT demonstration comprises (1) A sample sentence, (2) RPC for the sentence, and (3) The ground-truth sentence. These CoT demonstrations are integrated with the KNN-RPC-prompting inputs to serve as comprehensive prompting material for the LLM.}
    \label{fig:CoT KNN-Prompting}
\end{figure*}

\begin{figure*}[t]
    \centering
    \includegraphics[width=\textwidth]{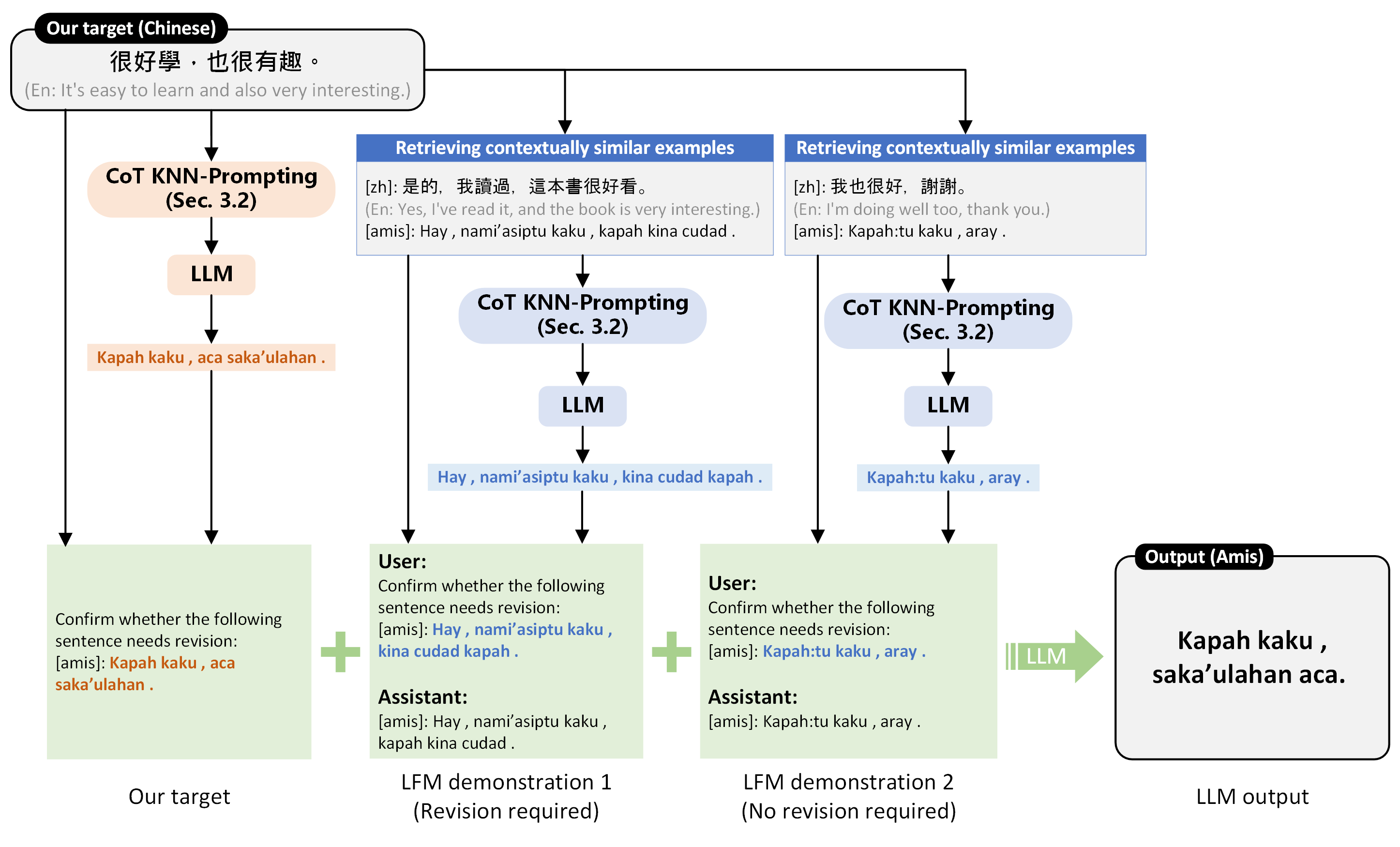}
    \caption{LFM Prompting}
    \label{fig:LFM Prompting}
\end{figure*}

\subsection{CoT Prompting}\label{Sec:CoT} In this methodology, we harness the CoT strategy, as delineated by Wei et al. \cite{wei2022chain}, to guide the LLM in effectively utilizing the RPC for translating a given sentence $s$. Specifically, this approach involves the following steps:

\begin{itemize}
    \item When presented with a sentence $s$ for translation, accompanied by KNN-RPC-prompting inputs (i.e., $k$ contextually similar examples and individual word translations), we further augment the LLM's input with $q$ CoT demonstrations. These demonstrations are designed to illustrate how to use the provided RPC to formulate the final translated sentences.

    \item An instance showcasing two CoT demonstrations is illustrated in Figure \ref{fig:CoT KNN-Prompting}. It is important to note that these CoT demonstrations are integrated with the KNN-RPC-prompting inputs to serve as comprehensive prompting material for the LLM.

    \item For detailed examples of this prompting structure, please refer to Table \ref{tab:Example-for-Simplified-CoT-KNN-Prompting} in the Appendix.

\end{itemize}

The overarching aim of this methodology is to empower the LLM with an understanding of the grammatical rules and the proficiency to fully leverage the RPC, including both the retrieved sentences and individual word translations, for producing coherent and accurate translations.

\subsection{Learning-from-Mistakes (LFM) Prompting}\label{sec:lfm}
LFM Prompting is a two-stage approach aimed at enhancing the quality of translations. This method leverages the result from the CoT KNN-Prompting and incorporates a feedback mechanism by conducting trial translation to refine the translation based on past translation errors. The method works in the following phases:

\begin{itemize} 
\item \textbf{Phase 1: Trial Translation with CoT Prompting}
When given a sentence $s$ to translate, we start by retrieving $q$ contextually similar sentence pairs from the data store. Each pair ($s_{q_i}, t_{q_i}$) consists of a Chinese sentence $s_{q_i}$ and its corresponding indigenous sentence $t_{q_i}$. For each $s_{q_i}$, we employ CoT KNN-Prompting (the method introduced in Section \ref{Sec:CoT}) to translate it, resulting in $\hat{t}_{q_i}$. At this stage, we have ($s_{q_i}$, $t_{q_i}$, $\hat{t}_{q_i}$). Our approach involves using these results as examples for the LLM to learn from its translation errors and make improvements.

\item \textbf{Phase 2: Learning from Past Mistakes}
The second phase of LFM Prompting introduces a crucial element: the incorporation of past translation errors. Specifically, we treat ($s_{q_i}$, $t_{q_i}$, $\hat{t}_{q_i}$) from the Phase 1 as LFM examples. In this phase, we present the LLM with a set of such examples, alongside the translation $\hat{t}$ generated by using CoT KNN-Prompting to translate $s$. The language model is tasked with refining $\hat{t}$ by considering the error examples in translation. It uses the provided examples of mistranslations to correct and improve the initial translation $\hat{t}$, aligning it more closely with the correct target language structure and meaning.
\end{itemize}

Furthermore, Figure \ref{fig:LFM Prompting} provides a visual representation of the entire architecture's workflow, illustrating the sequential processes outlined above. We also show a prompting example in Table \ref{tab:LFM}.

\input{Experiment}
\section{Conclusion}
\label{sec:conclusion}
This study delves into the capabilities of LLMs in translating indigenous languages. Despite a limited datastore of parallel translations, our introduced methodologies: \textit{KNN-Prompting with RPC, CoT Prompting, and LFM Prompting} demonstrate effectiveness in harnessing LLMs for this task. Emphasizing our technical contribution, empirical results highlight the superior performance of the CoT Prompting and LFM strategy over the compared baseline, signifying its adeptness at capturing intricate linguistic nuances and offering an advanced approach to preserving linguistic diversity.
\input{Limitations}

\section*{Acknowledgement}
This work is supported by NSTC 112-2634-F-005-002-project Smart Sustainable New Agriculture Research Center (SMARTer), NSTC Taiwan Project under grant 112-2221-E-005-075-MY3, and Ministry of Education, Taiwan. 

% \newpage
\bibliography{anthology,custom}
\bibliographystyle{acl_natbib}

% \newpage

\appendix
\input{appendix}

% \noindent\textbf{\large Appendix}

% \input{Dataset}
% \input{EvaluationMetrics}
% \input{ImplementationDetails}
\end{CJK*}
\end{document}

%% file: Experiment.tex
\begin{CJK*}{UTF8}{bsmi}
\section{Experiment}
\label{sec:experiment}

%南勢阿美族語
\begin{table*}[t]
\centering

\resizebox{.7\textwidth}{!}{
\begin{tabular}{l|cccc}
\hline
\multicolumn{5}{c}{\textbf{Southern Amis}} \\
\hline
\textbf{Methods}                                    & \textbf{BLEU1}$_{STD}$ & \textbf{BLEU2}$_{STD}$ & \textbf{BLEU3}$_{STD}$ & \textbf{chrF++}$_{STD}$ \\ \hline
\textbf{Zeroshot}                                   & 1.0          & 0.0               & 0.0 & 3.9 \\
\textbf{20-shots}                                   & 18.0          & 4.9          & 1.9 & 16.3 \\
\textbf{Knn-Prompting (k=5)}                        & 30.1          & 14.4          & 6.9 & 28.1 \\
\textbf{Knn-Prompting (k=10)}                       & 33.3          & 16.4          & 8.0 & 34.2 \\
\textbf{Knn-Prompting w. RPC (k=5)}  & 38.2$_{2.2}$          & 10.5$_{1.8}$          & 4.3$_{1.1}$ & 41.2$_{1.1}$ \\
\textbf{Knn-Prompting w. RPC (k=10)} & 37.8$_{2.2}$          & 12.5$_{3.2}$          & 5.2$_{1.9}$  & 41.5$_{1.6}$ \\
\textbf{CoT Prompting}            & 44.4$_{1.5}$          & 14.3$_{0.6}$          & 5.9$_{1.1}$  & 43.5$_{0.3}$ \\
% \textbf{CoT (q=2, k=10)}            & 44.6$_{2.5}$          & 18.0$_{1.6}$          & 8.5$_{1.0}$  & 45.1$_{1.8}$ \\
\textbf{LFM Prompting}    & 44.4$_{2.7}$          & 17.5$_{1.8}$          & 8.2$_{1.7}$ & 44.9$_{1.9}$ \\
\hline
\end{tabular}
}
\caption{\label{table:southern_amis} The translation results for \textbf{Southern Amis}}
\end{table*}

\begin{table*}[t]

\centering

\resizebox{.7\linewidth}{!}{
\begin{tabular}{l|cccc}
\hline
\multicolumn{5}{c}{\textbf{Coastal Amis}} \\ % Title for the first table
\hline
\textbf{Methods}                                    & \textbf{BLEU1}$_{STD}$ & \textbf{BLEU2}$_{STD}$ & \textbf{BLEU3}$_{STD}$ & \textbf{chrF++}$_{STD}$  \\ \hline
\textbf{Knn-Prompting w. RPC (k=5)}    & 42.9$_{1.8}$ & 11.8$_{0.9}$ & 4.7$_{1.1}$ & 45.4$_{0.8}$   \\
\textbf{Knn-Prompting w. RPC (k=10)}   & 43.3$_{1.2}$ & 13.4$_{0.6}$ & 5.8$_{0.8}$ & 44.8$_{1.1}$   \\
\textbf{CoT Prompting}  & 44.5$_{2.8}$ & 11.9$_{3.0}$ & 4.7$_{2.3}$ & 45.7$_{1.6}$   \\
% \textbf{CoT (q=2, k=10)} & 44.6$_{2.4}$ & 13.0$_{2.9}$ & 5.2$_{2.4}$ & 46.3$_{2.0}$   \\
\textbf{LFM Prompting} & 44.1$_{2.0}$ & 12.6$_{2.9}$ & 5.7$_{2.5}$ & 46.1$_{1.8}$  \\
\hline
\end{tabular}}\\
\resizebox{.7\linewidth}{!}{
\begin{tabular}{l|cccc}
\hline
\multicolumn{5}{c}{\textbf{Wanda Tayal}} \\ % Title for the second table
\hline
\textbf{Methods}                                    & \textbf{BLEU1}$_{STD}$ & \textbf{BLEU2}$_{STD}$ & \textbf{BLEU3}$_{STD}$ & \textbf{chrF++}$_{STD}$  \\ \hline
\textbf{Knn-Prompting w. RPC (k=5)}    & 41.5$_{2.5}$ & 13.0$_{2.1}$ & 4.8$_{1.6}$ & 42.5$_{2.4}$  \\
\textbf{Knn-Prompting w. RPC (k=10)}   & 42.1$_{2.2}$ & 13.6$_{2.7}$ & 5.7$_{1.4}$ & 42.8$_{2.6}$  \\
\textbf{CoT Prompting}  & 46.3$_{1.8}$ & 14.4$_{2.6}$ & 5.8$_{1.2}$ & 44.7$_{2.1}$  \\
% \textbf{CoT (q=2, k=10)} & 46.3$_{1.8}$ & 14.7$_{2.6}$ & 6.0$_{1.2}$ & 44.4$_{2.2}$  \\
\textbf{LFM Prompting} & 45.2$_{2.0}$ & 14.0$_{1.7}$ & 5.8$_{0.9}$ & 43.9$_{2.0}$    \\
\hline
\end{tabular}}\\

\resizebox{.7\linewidth}{!}{
\begin{tabular}{l|cccc}
\hline
\multicolumn{5}{c}{\textbf{Siji Tayal}} \\ % Add this line for the new row
\hline
\textbf{Methods}                                    & \textbf{BLEU1}$_{STD}$ & \textbf{BLEU2}$_{STD}$ & \textbf{BLEU3}$_{STD}$ & \textbf{chrF++}$_{STD}$  \\ \hline
\textbf{Knn-Prompting w. RPC (k=5)}    & 44.3$_{3.2}$ & 14.6$_{2.1}$ & 4.9$_{1.7}$ & 39.3$_{2.0}$  \\
\textbf{Knn-Prompting w. RPC (k=10)}   & 44.4$_{3.0}$  & 14.5$_{2.0}$ & 5.4$_{2.3}$ & 40.9$_{1.8}$  \\
\textbf{CoT Prompting}  & 47.5$_{2.7}$ & 16.0$_{1.2}$ & 5.9$_{1.4}$ & 41.2$_{1.0}$   \\
% \textbf{CoT (q=2, k=10)} & 49.2$_{1.4}$ & 18.8$_{1.7}$ & 8.2$_{2.1}$ & 43.5$_{1.6}$   \\
\textbf{LFM Prompting}      & 50.0$_{1.2}$ & 20.0$_{1.4}$ & 9.3$_{2.0}$ & 43.4$_{2.0}$  \\
\hline
\end{tabular}
}
\resizebox{.7\linewidth}{!}{
\begin{tabular}{l|cccc}
\hline
\multicolumn{5}{c}{\textbf{Duda Seediq}} \\ % Title for the first table
\hline
\textbf{Methods}                                    & \textbf{BLEU1}$_{STD}$ & \textbf{BLEU2}$_{STD}$ & \textbf{BLEU3}$_{STD}$ & \textbf{chrF++}$_{STD}$  \\ \hline
\textbf{Knn-Prompting w. RPC (k=5)}    & 45.0$_{1.2}$ & 16.2$_{1.5}$ & 5.4$_{0.8}$ & 38.2$_{0.8}$   \\
\textbf{Knn-Prompting w. RPC (k=10)}   & 45.7$_{1.2}$ & 17.1$_{1.4}$ & 6.7$_{1.5}$ & 39.3$_{1.6}$   \\
\textbf{CoT Prompting}  & 46.1$_{1.6}$ & 17.5$_{1.4}$ & 6.9$_{1.0}$ & 38.9$_{1.1}$   \\
% \textbf{CoT (q=2, k=10)} & 46.2$_{1.6}$ & 17.6$_{1.4}$ & 6.4$_{1.1}$ & 39.2$_{1.2}$   \\
\textbf{LFM Prompting} & 46.3$_{1.5}$ & 17.3$_{2.1}$ & 6.9$_{1.4}$ & 39.3$_{1.2}$  \\
\hline
\end{tabular}}
\caption{\label{table:coastal_amis} The translation results for \textbf{Coastal Amis}, \textbf{Wanda Tayal}, \textbf{Siji Tayal}, and \textbf{Duda Seediq}}
\end{table*}

\subsection{Model Usage}
Utilizing the GPT-3.5-turbo-16k-0613 version with a temperature setting of 0, we employ Sentence BERT \cite{reimers2019sentence} as the embedding model to retrieve $k$-nearest neighbor sentences. The similarity between sentences is computed using cosine similarity.

\subsection{Data Sets}
We use the learning materials for various indigenous languages from the 'Klokah' website \footnote{\url{https://web.klokah.tw/}} provided by the Foundation for the Research and Development of Indigenous Languages in Taiwan as our evaluation corpora. Each indigenous group consists of 450 sentences with corresponding Chinese translations and a dictionary of 1000 words (single word translation). For each language, we divide this dataset into two parts: 
\begin{itemize}
    \item \textbf{Test Data - } A random selection of 100 sentences was used to evaluate the translation performance of various methods. 
    \item\textbf{Reference Data - } The remaining 350 sentences and all dictionaries were used as reference materials for the LLM translation.
\end{itemize}

\subsection{Evaluation Results}
\subsubsection{Automatic Score}
We've employed the GPT-3.5-turbo as our foundational language model for translation tests. Initially, we opted for Southern Amis, an indigenous language, as our primary focus, evaluating translation accuracy using the standard BLEU \cite{papineni2002bleu} and chrF++ \cite{popovic2017chrf++} metrics. As depicted in Table \ref{table:southern_amis}, the zero-shot translation results indicate the model's limitations in effectively translating this language in the absence of reference data, reflected in BLEU scores nearing zero. However, introducing 20-shot reference data prompts the model to engage in \cite{agrawal2022context} in-context learning, resulting in a marginal improvement in BLEU scores. This highlights the potential of few-shot learning.

Furthermore, from the results in Table \ref{table:southern_amis}, we can observe that using KNN-Prompting by retrieving contextual-relevant examples improves translation quality. We can also observe that utilizing the Chain-of-Thought strategy to guide the LLM also brings an improvement in translation quality, with an increase in BLEU scores from 1 to 3. We also report experiment results with Coastal Amis, Wanda Tayal, Siji Tayal, and Duda Seediq languages. The results are shown in Tables \ref{table:coastal_amis}. 

%海岸阿美族語

When comparing models, we use BLEU3 as the main performance metric, as BLEU3 considers 3-gram matches, offering a more holistic view of the quality of translations, particularly in terms of fluency and coherence. In terms of BLEU3 scores, CoT Prompting consistently surpasses the base KNN-Prompting across all languages. This points towards the importance of capturing longer sequences and understanding the grammatical flow of the indigenous languages. We can also see the performance boost when we employ the LFM strategy with CoT in the compared languages.

\subsubsection{Qualitive Review by Language Expert}

In Table \ref{tab:my_label_expert} in the appendix, we present the results of an evaluation by a Coastal Amis language expert, assessing translations from Chinese into Coastal Amis using various methods. This offers insights into the effectiveness of these translation strategies and helps us understand their impact on translation quality.

\begin{itemize}
    \item Initially, the expert demonstrates a preference for translations produced by the LFM method, highlighting its contribution to linguistic precision and affirming the significant role of the LFM phase in enhancing translation quality.
    
    \item In the second dialogue, the expert's endorsement of the COT and LFM method suggests that its incorporation can refine the LLM's understanding and conveyance of the target language's nuances.
    
    \item There is an identified need for improvement in translating sentence structures, particularly with time adverbs such as "非常" (very), "很" (very), "最" (most), and the placement of temporal terms like "今天" (today), "明天" (tomorrow) at the sentence's end. The LFM method is anticipated to guide the LLM in learning and internalizing these linguistic patterns, thereby refining the translations. Challenges such as dictionary absences are addressed by seeking synonyms, for instance, substituting "專為" (specially designed for) with "最" (most), and "南邊" (south) with "藍色" (blue). We posit that expanding the dictionary will mitigate such issues, further enhancing translation fidelity.
\end{itemize}

Overall, the expert's reviews imply that the translation approach integrating the LFM strategy tends to yield more precise and culturally attuned translations. This suggests that for LLMs translating less-resourced languages, a strategy amalgamating error feedback with accumulative learning might prove more effective. These insights bolster the methodologies delineated in our paper, positing that a stratified and iterative enhancement approach can substantially uplift translation quality, particularly for languages with constrained structural and lexical resources.

% We have invited experts in Coastal Amis language to review our translation results. We have also identified some rules, documented in the description of Table \ref{tab:my_label_expert}. 

% While formulating detailed rules based on the grammar of indigenous languages or annotating corpora may be a feasible approach, the scarcity of these language experts poses a significant challenge, making the process complex and inefficient. Therefore, our goal is to enhance translation performance effectively, particularly in the aspects of structure and specific rules, by adopting LFM. This allows LLM to leverage reasoning abilities while relying on limited expert feedback to achieve this objective. 
% 改成以下這段？
% 我們已邀請海岸阿美族語言專家來審查我們的翻譯結果。而我們也有在其中發現了一寫規則，寫在了Table 4 的敘述當中。雖然制定基於土著語言語法的詳盡規則或對語料進行標註可能是一種可行的方法，但這些語言專家的短缺帶來了重大挑戰，使得這個過程變得複雜且低效。因此，我們的目標是透過採用（LFM）有效提升翻譯表現，特別是在結構和具體規則方面。這使得大型語言模型（LLM）能夠充分發揮推理能力，同時倚賴有限的專家回饋來達到這個目的。
% While defining a detailed set of rules or annotating corpora based on the grammar of indigenous languages might be an approach, the scarcity of experts in indigenous languages poses a significant challenge, making it both difficult and inefficient. Therefore, we aim to enhance translation performance effectively by utilizing LFM , allowing LLM to leverage reasoning abilities in conjunction with limited expert feedback. The information gathered in Table \ref{tab:my_label_expert} represents what we aim for LLM to learn through the LFM method, encompassing structures and specific rules.
\begin{table*}[t]
    \centering
 \begin{tabular}{p{14cm}}
\toprule
\textbf{Coastal Amis language expert reviews translation results} \\
\midrule
% Please analyze the differences between [your answer] and [Revision] results. \\
\texttt{[zh]}:你喜歡看書嗎？ (En: Do you like reading books?) \\
\texttt{[CoT]}: Kiso (你, you) maolah (喜歡, like) minengneng (看望, visit) a cudad (書, book) haw ? \\
\texttt{[LFM]}: Kiso (你, you) maolah (喜歡, like) cudad (書, book) minengneng (看望, visit) haw ? \\
\texttt{[language expert]}: Maolah (喜歡, like) kiso (你, you) minengneng (看望, visit) to codad (書, book) haw? \\
\midrule 
\texttt{[zh]}:是的，我很喜歡。(En: Yes, I really like it.)\\
\texttt{[CoT]}: Hai (是的, yes) , maolah (很喜歡, really like) to kako (我, I) . \\
\texttt{[LFM]}: Hai (是的, yes) , maolah (很喜歡, really like) to kako (我, I) . \\
\texttt{[language expert]}: Hai (是的, yes), ma`olah (很喜歡, really like) kako (我, I). \\
\midrule
\texttt{[zh]}:你會織布嗎？(En: Do you know how to weave?)\\
\texttt{[CoT]}: Mafana` (會, know how to) kiso (你, you) a miteno`oy (織布, Weaving fabric) ?\\
\texttt{[LFM]}: Mafana` (會, know how to) kiso (你, you) miteno`oy (織布, Weaving fabric) ?\\
\texttt{[language expert]}: Mafana` (會, know how to) kiso (你, you) miteno`oy (織布, Weaving fabric) haw? \\
\midrule 
\texttt{[zh]}:我最喜歡夏天。(En: I like summer the most.)\\
\texttt{[CoT]}: Maolah  (很喜歡, really like) kako (我, I) patodongan (專為, Specially designed for) ko kaciherangan  (夏天, summer) .\\
\texttt{[LFM]}:Maolah  (很喜歡, really like) kako (我, I) kaciherangan  (夏天, summer) patodongan (專為, Specially designed for) .\\
\texttt{[language expert]}: O kaciherangan (夏天, summer) ko kaolahan (最喜愛的, favorite) ako(我, I). \\
\midrule
\texttt{[zh]}:我最喜歡藍色。(En: I like blue the most.)\\
\texttt{[CoT]}: Maolah (喜歡, like) kako (我, I) patodongan (專為, Specially designed for) satimolan (南邊, south) .\\
\texttt{[LFM]}: Maolah (喜歡, like) kako (我, I) satimolan (南邊, south) patodongan (專為, Specially designed for) .\\
\texttt{[language expert]}: O langdaway (藍色, blue) a cengel (顏色, color) ko kaolahan (最喜歡, favorite) ako (我, I). \\
\midrule
\texttt{[zh]}:是的，天氣非常好。(En: Yes, the weather is very nice.)\\
\texttt{[CoT]}: Hai (是的, yes) , romi`ad (天氣, weather) tada (非常, very) nga`ay (好, good) .\\
\texttt{[LFM]}: Hai (是的, yes) , romi`ad (天氣, weather) tada (非常, very) nga`ay (好, good) .\\
\texttt{[language expert]}: Hai (是的, yes), fangcal (很好, very good) ko romi`ad (天氣, weather) anini (今天, today).\\

\bottomrule
\end{tabular}
    \caption{Result of Language Expert Review}
    \label{tab:my_label_expert}
\end{table*}
\end{CJK*}

%% file: Limitations.tex
\section{Limitations}
The strength of this framework lies in its capacity to translate less common, niche languages with a limited number of examples. Nevertheless, several challenges were encountered during the experiments. For example, in the case of the Southern Amis language, the term `we` can be translated as `kami` or `niyam,` among other options. Determining whether these terms carry subtle distinctions in meaning or are interchangeable necessitates the expertise of native speakers. Moreover, the use of the BLEU metric provides only one standardized answer, which may not consistently align with the actual context.

Furthermore, within the LFM context, structural or grammatical corrections are solely guided by prior examples, as the language model itself lacks the capability for independent reasoning and adjustment. Therefore, achieving significant breakthroughs in effectiveness remains a challenge. Finally, while our methods have demonstrated data-driven enhancements, they do not fully address the issue of insufficient few-shot data resulting in inconsistent translation outcomes. Further research and innovation are essential in addressing this matter.

%% file: appendix.tex
\section*{Appendix}

\begin{table*}[]
    \centering
 \begin{tabular}{p{14cm}}
\toprule
\textbf{Knn-Prompting with RPC} \\
\midrule
You are an Amis language translator. The followings are some [zh] to [amis] examples.\\
Chinese: 是的，我讀過，這本書很好看。(English: Yes, I've read it, and the book is very interesting.)\\
\texttt{[Amis]}: Hay, nami’asiptu kaku, kapah kina cudad. \\
\texttt{[zh]}: 我也很好，謝謝。(English: I'm doing well too, thank you.)\\
\texttt{[Amis]}: Kapah:tu kaku, aray. \\
\texttt{[zh]}: 郵差也感到喜悅與滿足。(English: The postman also feels joy and satisfaction.)\\
\texttt{[Amis]}: U yu-cay satu, mikihatiya a lipahak, a mi’edem tu ulah nu valucu’. \\
~~~~: \\
~~~~: \\
\texttt{[zh]}: 很好學 (English: Easy to learn.) \\
\texttt{[Amis]}: kapah \\
\texttt{[zh]}: 也 (English: Also) \\
\texttt{[Amis]}: aca \\
\texttt{[zh]}: 很有趣 (English: Very interesting).\\
\texttt{[Amis]}: saka’ulahan \\
Based on the above examples. Could you help to translate [zh]: 很好學，也很有趣. (English: It's easy to learn and interesting)\\
\bottomrule
\end{tabular}
    \caption{Simplified Example for Knn-Prompting with RPC}
    \label{tab:RPC}
\end{table*}

\begin{table*}[]
    \centering\small  
 \begin{tabular}{p{14cm}}
\toprule
\textbf{CoT Demonstration 1} \\
\midrule
You are an Amis language translator. The followings some [zh] to [amis] examples.\\
\texttt{[zh]}: 現在幾點鐘？ (En: What time is it now?)\\
\texttt{[Amis]}: Pina’ay ku tuki anini ? \\
\texttt{[zh]}: 喔！她什麼時候回來呢？(En: Oh! When is she coming back?)\\
\texttt{[Amis]}: A, a hacuwa cira a taluma’ ? \\
~~~~: \\
~~~~: \\
\texttt{[zh]}: 要 (En: Need)\\
\texttt{[Amis]}: aw \\
\texttt{[zh]}: 走 (En: Go)\\
\texttt{[Amis]}: rakat \\
% \texttt{[*zh]}: 多久 (En: How long.)\\
\texttt{[zh]}: 幾天 (En: How many days.)\\
\texttt{[Amis]}: kapina a remi’ad \\
\texttt{[zh]}: 呢 (En: Question particle)\\
\texttt{[Amis]}: saw \\
Based on the above examples. Could you help to translate [zh]: 要走多久呢？\\
\texttt{[Assistant:]} Hacuwa ku tenes a remakat ?  \\
\end{tabular}
\begin{tabular}{p{14cm}}
\toprule
\textbf{CoT Demonstration 2} \\
\midrule
You are an Amis language translator. The followings some [zh] to [amis] examples.\\
\texttt{[zh]}: 今年，我們伯伯全家人從台北搭車回來。(En: This year, our uncle's entire family came back from Taipei by car.)\\
\texttt{[Amis]}: Anini a miheca, makakarireng a taluma’ ku vaki niyam atu wawa nira namaka Taypey. \\
\texttt{[zh]}: 火車比較快。(En: The train is faster.)\\
\texttt{[Amis]}: U silamalay ku kalamkamay. \\
~~~~: \\
~~~~: \\
\texttt{[zh]}: 要 (En: Need)\\
\texttt{[Amis]}: aw \\
\texttt{[zh]}: 搭什麼 (En: Take what.)\\
\texttt{[Amis]}: Makama’an \\
% \texttt{[*zh]}: 車 (En: Car)\\
\texttt{[zh]}: 公車 (En: Bus)\\
\texttt{[Amis]}: vasu \\
\texttt{[zh]}: 到 (En: Arrive)\\
\texttt{[Amis]}: tangasa \\
\texttt{[zh]}: 台北 (En: Taipei)\\
\texttt{[Amis]}: Taypak \\
Based on the above examples. Could you help to translate [zh]: 要搭什麼（車）到台北？\\
\texttt{[Assistant:]} Makama’an a tala i Taypak ? \\
\end{tabular}
 \begin{tabular}{p{14cm}}
\toprule
\textbf{CoT Prompting} \\
\midrule
You are an Amis language translator. The followings are some [zh] to [amis] examples.\\
Chinese: 是的，我讀過，這本書很好看。(English: Yes, I've read it, and the book is very interesting.)\\
\texttt{[Amis]}: Hay, nami’asiptu kaku, kapah kina cudad. \\
\texttt{[zh]}: 我也很好，謝謝。(English: I'm doing well too, thank you.)\\
\texttt{[Amis]}: Kapah:tu kaku, aray. \\
\texttt{[zh]}: 郵差也感到喜悅與滿足。(English: The postman also feels joy and satisfaction.)\\
\texttt{[Amis]}: U yu-cay satu, mikihatiya a lipahak, a mi’edem tu ulah nu valucu’. \\
\texttt{[zh]}: 很好學 (English: Easy to learn.) \\
\texttt{[Amis]}: kapah \\
\texttt{[zh]}: 也 (English: Also) \\
\texttt{[Amis]}: aca \\
\texttt{[zh]}: 很有趣 (English: Very interesting).\\
\texttt{[Amis]}: saka’ulahan \\
Based on the above examples. Could you help to translate [zh]: 很好學，也很有趣. (English: It's easy to learn and interesting)\\
\bottomrule
\end{tabular}
    \caption{Example for Simplified CoT KNN-Prompting}
    \label{tab:Example-for-Simplified-CoT-KNN-Prompting}
\end{table*}

\begin{table*}[]
    \centering
 \begin{tabular}{p{14cm}}
\toprule
\textbf{LFM Prompting Example} \\
\midrule
% Please analyze the differences between [your answer] and [Revision] results. \\
Please analyze the differences between \texttt{[Your Answer]} and \texttt{[Correct Answer]} results. \\
\texttt{[zh]}:是的，我讀過，這本書很好看。 (En: Yes, I've read it, and the book is very interesting.) \\
\texttt{[Your Answer]}:Hay , nami’asiptu kaku , kina cudad kapah .\\
\texttt{[Correct Answer]}: Hay , nami’asiptu kaku , kapah kina cudad . \\
Please analyze the differences between \texttt{[Your Answer]} and \texttt{[Correct Answer]} results. \\
\texttt{[zh]}:我也很好，謝謝。 (En: I'm doing well too, thank you.) \\
\texttt{[Your Answer]}:Kapah:tu kaku , aray . \\
\texttt{[Correct Answer]}: Kapah:tu kaku , aray . \\
\midrule 
You are an Amis language translator. The followings some [zh] to [amis] examples.\\
\texttt{[zh]}:是的，我讀過，這本書很好看。(En: Yes, I've read it, and the book is very interesting.)\\
\texttt{[amis]}:Hay , nami’asiptu kaku , kapah kina cudad . \\
\texttt{[zh]}:我也很好，謝謝。(En: I'm doing well too, thank you.)\\
\texttt{[amis]}:Kapah:tu kaku , aray . \\
\texttt{[zh]}:郵差也感到喜悅與滿足。 (En: The postman also feels joy and satisfaction.)\\
\texttt{[amis]}:U yu-cay satu , mikihatiya a lipahak , a mi’edem tu ulah nu valucu’ . \\
...\\
\texttt{[zh]}: 很好學 (En: Easy to learn.)\\
\texttt{[amis]}: kapah\\
\texttt{[zh]}: 也 (En: Also)\\
\texttt{[amis]}: aca\\
\texttt{[zh]}: 很有趣 (En: Very interesting.)\\
\texttt{[amis]}: saka’ulahan\\
Check whether the following sentence needs revision: \\\texttt{[zh]}:很好學，也很有趣。 (English: It's easy to learn and interesting)
\\\texttt{[Your Answer]}:Kapah kaku , aca saka’ulahan . \\
\texttt{[Correct Answer]}:\\
\bottomrule
\end{tabular}
    \caption{Example for Simplified LFM Prompting. Note that as introduced in the LFM method, when given a sentence $s$ (i.e., 很好學，也很有趣) to translate, we start by retrieving $q$ contextually similar sentence pairs from the data store and use CoT KNN prompting to obtain trial translation results (the sentence followed \texttt{[Your Answer]}) and also the correct answer for enabling LFM.}
    \label{tab:LFM}
\end{table*}